\documentclass[twoside]{article}

\usepackage{subcaption}
\usepackage[inkscapelatex=false]{svg}
\usepackage{graphicx}

\usepackage{authblk}

\def\authorEmail{sergio.ramirez@thalesaleniaspace.com}

\author[1]{Sergio Ram\'irez-Gallego\thanks{Corresponding author. E-Mail: \authorEmail}}

\affil[1]{Thales Alenia Space Spain, Tres Cantos, Spain}

\title{On-board Remote-Sensing Foundation Models for Unsupervised Change Detection of Disaster Events}

\begin{document}
\maketitle

\begin{abstract}
Remote Sensing Foundation Models (RSFMs) have emerged as a powerful alternative to supervised models for Earth Observation, allowing satellites to autonomously trigger high-resolution captures or adjust tasking parameters upon detecting an anomaly, thereby maximizing the utility of the mission's limited power and computational resources. RSFMs are versatile, unified encoders that optimize onboard storage for multiple orbital applications while ensuring high-fidelity feature extraction. 

In particular, unsupervised change detection with RSFMs offers a well-informed and transformative path for disaster monitoring without expensive labels. In this paper, we present a novel unsupervised detection method based on ResNet (RSFM) + FPN which identifies a wide spectrum of anomalies by detecting subtle semantic shifts in the latent space between successive orbital passes. 

By relying on an untrained FPN architecture and its intrinsic priors, the system achieves efficient image-level generation and higher resolution mapping with minimal effort (training-free) compared to previous proposals (patch-based, trained). 
And by replacing tailored models with RSFMs, we can achieve comparable results through an approach that eliminates the need for bespoke training and extensive development effort and adds customization, while ensuring high-performance generalization across diverse terrains and sensors.
\end{abstract}

\section{Introduction}

The pursuit of on-board Machine Learning (ML) for Earth Observation is driven by the fundamental need to transform satellites from passive collectors into autonomous decision-makers~\cite{furano20, duggan25}. By moving the analytical "intelligence" to the orbital edge, satellites can perform real-time data reduction, significantly mitigating the "data deluge" that currently overwhelms ground-station downlinks. This shift is particularly compelling for time-sensitive applications; instead of transmitting gigabytes of redundant imagery, an on-board model can extract and prioritize only the most salient information, such as, the precise coordinates of a wildfire front or the extent of a sudden flood. This means sending low-bandwidth alerts that reach emergency responders in minutes instead of hours~\cite{vit22}. Furthermore, on-board ML enables a more resilient and adaptive sensing strategy, allowing the satellite to autonomously trigger high-resolution captures or adjust tasking parameters upon detecting an anomaly, thereby maximizing the utility of the mission’s limited power and computational resources.

Unsupervised change detection offers a transformative path for disaster monitoring by eliminating the dependence on exhaustive, human-labeled datasets which are often unavailable for rare or localized extreme events~\cite{vit22}. Unlike supervised models that are confined to recognizing specific, pre-defined classes like smoke or water, unsupervised models can identify and prioritize a wide spectrum of anomalies by detecting subtle semantic shifts in the latent space between successive orbital passes. 

Although the development and posterior deployment of lightweight AI models have significantly advanced the field of orbital inference~\cite{sang26, smith25}, "small" models often suffer from a "generalization gap" that limits their efficacy amidst the volatile and diverse conditions inherent in global Earth observation. RSFMs~\cite{stewart23} have emerged as a powerful alternative, offering robust, pre-trained representations that demonstrate exceptional adaptability across varying geographic regions and sensor modalities. Nevertheless, transitioning large-scale models from ground-based servers to the orbital edge—specifically onto resource-constrained platforms (i.e.: nanosatellites) presents a formidable engineering hurdle~\cite{sang26}. Overcoming stringent limitations in onboard memory, power budgets, and computational throughput, alongside the necessity for radiation-hardened reliability, remains a primary obstacle to realizing the full potential of RSFMs in space.

In this work, we propose~\textit{Unsupervised Detection Feature Pyramid Network (UDFPN)}, a method that provides a robust and efficient framework for autonomous disaster detection. By combining a self-supervised ResNet backbone~\cite{stewart23} with an untrained FPN~\cite{lin17}, UDFPN overcomes the common problem of spatial degradation in deep networks~\cite{he2015deepresiduallearningimage}. This architecture allows the model to produce semantically rich and geometrically sharp embeddings without extra training. Our results demonstrate that UDFPN significantly outperforms similar benchmarks (i.e.: PANN) in landslide detection and generates highly granular change maps for complex events.
Unlike patch-based alternatives that require dozens of repetitive inferences, our image-level approach proposes single forward passes with emphasis in matrix optimization. Furthermore, UDFPN offers a flexible, compressed architecture that can be tailored to the specific resource limits of a mission. 

We advocate for the transition from tailored models to RSFMs, which significantly reduce the engineering overhead associated with model training and provide a wide range of advantages. RSFMs provide a unified backbone capable of generalizing across varied terrains and disparate sensor inputs without the need for bespoke development. This work serves as the initial phase in our transition toward that objective.

\section{Methodology}

\subsection{Data}

To evaluate the robustness of the UDFPN, a dataset was derived using Landsat-8 Operational Land Imager (OLI) data from~\cite{smith_25_data}. The resulting selection comprised eight events: two fires, two floods, and four landslides. We utilized the eleven Landsat-8 Top of Atmosphere (TOA) bands featuring a 30 m spatial resolution (see Section~\ref{subsec:preprocess}). For the temporal sequence, original authors enforced a maximum cloud-cover threshold of 40\% per acquisition. Each event consists of a five-image time series (four pre-event, one post-event). Only cloud masks were considered for post-event instances in order to compute more realistic metrics. Data pipeline is replicated from~\cite{smith25, smith_25_data}.

\subsection{Preprocessing}
\label{subsec:preprocess}
Landsat-8 imagery was standardized into an 11-channel tensor format to ensure spectral consistency. Available bands were mapped to their respective spectral indices, with missing channels (e.g., Coastal Aerosol and Cirrus) initialized as zero-padding. To handle missing data, NaN values in the reflective bands were imputed with $0.0$.
All channels were then normalized to a $[0, 1]$ range using a Min-Max shift based on histograms. Finally, embeddings were L2-norm for computing the novelty score along the channel dimension.

\subsection{Model}

\begin{figure}[t]
    \centering
    \includegraphics[width=.95\columnwidth]{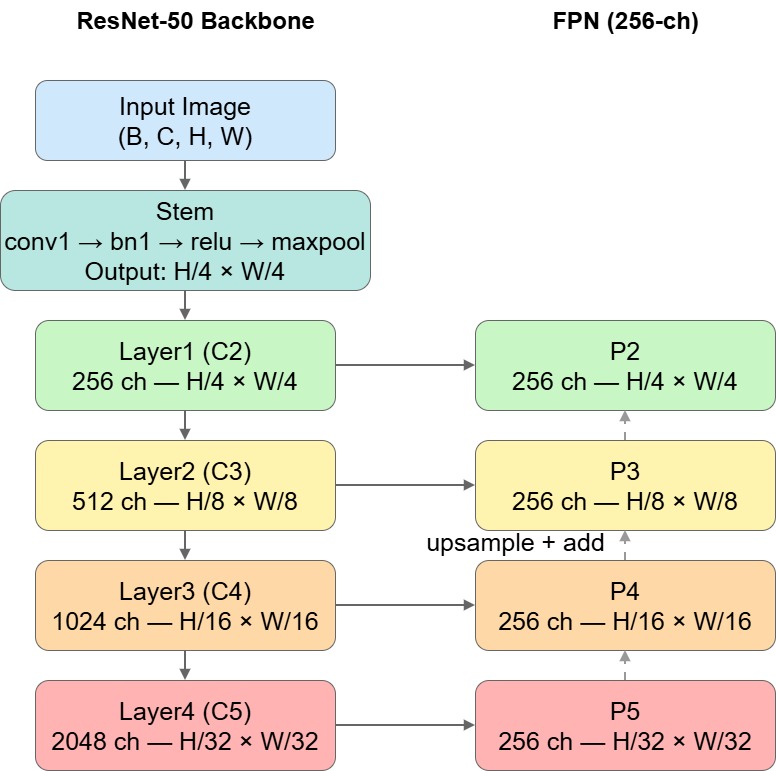}
	\caption{UDFPN architecture with ResNet-50 + FPN; output dimension = 256}
	\label{fig:architect}
\end{figure}

For robust feature extraction, we utilize a ResNet-X~\cite{he2015deepresiduallearningimage} (default: ResNet-50) backbone initialized with self-supervised weights from the SSL4EO-L~\cite{stewart23} dataset. This ensures the model is inherently tuned to the spectral signatures and spatial characteristics of satellite imagery, providing a superior foundation for change detection. See Figure~\ref{fig:architect} for a complete model architecture description.

Direct application of backbone embeddings is not feasible for unsupervised change detection at image level so an important architecture tune was made by integrating a multi-scale feature fusion network (FPN). In FPN, lower-resolution feature maps are spatially doubled using nearest-neighbor upsampling and integrated with their corresponding bottom-up counterparts. This integration is achieved through element-wise addition after the bottom-up maps have undergone $1 \times 1$ convolutions to ensure channel dimension alignment.

In ResNet-50, the loss of spatial meaning in the last layer is driven by a 32x cumulative downsampling factor and the exponential expansion of the effective receptive field across successive residual stages. 
This mechanism enforces spatial invariance which is not appropriate for generating spatially meaningful embeddings. 


To overcome the previous problem, UDFPN employs an untrained, frozen FPN as a structural aggregator~\cite{lin17}. By leveraging architectural inductive bias~\cite{uly20}, this process re-aligns abstract backbone features into a spatially coherent grid, producing embeddings that are both semantically rich and geometrically sharp without the need for additional training. 


To prioritize high-resolution spatial recovery, UDFPN extracts the final output from the second level (derived from the $C_2$ backbone block) of the pyramid connected to ResNet-50. The resulting output is a spatially dense embedding tensor with the following dimensions:
\begin{itemize}
    \item Height \& Width: $H/4 \times W/4$ of the original input resolution, preserving sufficient detail for localized change mapping.
    \item Embedding Dimension: 256 channels (channel output).
\end{itemize}

This architecture schema yields a final embedding representation of size $(256, H/4, W/4)$, where each embedding vector encodes a complex local signature used to compute detection distance between pre- and post-event imagery.

\subsection{Change detection score}

The proposed change detection metric quantifies localized landscape variations by computing the minimum cosine distance between a post-event query vector and a local search neighborhood within the pre-event embedding manifold. This approach evaluates change as a function of semantic displacement rather than raw intensity variance.

For a given pixel at coordinates $(i, j)$ in the post-event feature map, the change score $S_{i,j}$ is defined as the distance between the query embedding $e_{post}$ and its most semantically similar counterpart $e_{pre}$ within a defined spatial window in the pre-event map. 

This methodology is particularly robust for remote sensing applications where atmospheric conditions or sensor angles might change between acquisitions as we found in the reduced set of events evaluated. We have tested other scores more tailored to relationships between pre-events, such as: Z-score, distance to median/average, etc. These metrics though showed low effectiveness, probably due to pre-event images being dominated by clouds or low-light conditions. 

\subsection{Compression}

In contrast to fixed embeddings~\cite{smith25, vit22}, our method allows custom compression sizes for every axis. Dimensions are controlled by FPN output channels (default: $256$), while spatial extension of embeddings is controlled by the selected FPN output layer (downsampling factors: $[4, 8, 16, 32]$). This means that depending on the space mission's requirements (storage space and computing power), embeddings' shape could be adjusted conveniently. 

\section{Results}
\label{sec:results}

Output change maps were generated for the entire evaluation dataset~\cite{smith_25_data}. To assess detection performance, generated maps were leveraged alongside ground-truth labels to derive the \textit{Area Under the Precision-Recall Curve (AUPRC)}. PR curves were computed considering all spatial axis of embeddings and contrasting them to ground truth maps created by~\cite{vit22}. To slightly maintain data integrity and reduce noise, pixels containing cloud cover in the post-event acquisition were not considered. Note that ground truth maps are 1:1 with input images so they are downsized to match the spatial dimensions in the embedding space. In our case, x4 downsampling factor, while 16x for PANN~\cite{smith25}. This means fair comparisons among methods are not direct. Note also that PANN follows a patch-level generation strategy. 

Despite some notable differences in experiments and methodologies, the closest comparison to UDFPN is PANN~\cite{smith25} as it was evaluated on the same Landsat data. Table~\ref{table:auprc} shows metric-based outcomes between our method and PANN for the Landsat-8 subset~\cite{smith_25_data}. Experiments point to reasonable performance of our alternative, with special relevance for landslide events where our approach is better with a clear advantage. Concerning floods, we notice that one (out of two) event obtains quite low performance. The post-event image here was primarily characterized by variations in the river’s spectral properties. The model appears to have a detection bias toward structural changes, potentially leading to lower sensitivity for events characterized solely by water color variation.

\begin{table}[h]
    \centering
    \begin{tabular}{|l|c|c|c|}
        \hline
        \textbf{Model} & \textbf{Fires} & \textbf{Landslides} & \textbf{Floods} \\ \hline
        UDFPN & 71.38 $\pm$ 0.2 & \textbf{73.08} $\pm$ 0.4 & 32.99 $\pm$ 0.2 \\ \hline
        UDFPN-18 & 75.13 $\pm$ 0.3 & 39.90 $\pm$ 0.1 & 43.92 $\pm$ 0.1 \\ \hline
        PANN              & \textbf{86.07} $\pm$ 0.02 & 53.3 $\pm$ 0.2 & \textbf{53.9} $\pm$ 0.1 \\ \hline
    \end{tabular}
    \caption{Comparison of change detection performance (AUPRC) across different event types.}
\label{table:auprc}
\end{table}

\begin{figure}[htbp]
     \centering
     \begin{subfigure}[b]{0.24\textwidth}
         \centering
         \includegraphics[width=\textwidth]{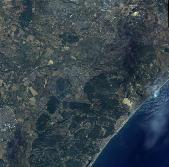}
         \caption{Pre-event}
         \label{fig:img1}
     \end{subfigure}
     \hfill 
     \begin{subfigure}[b]{0.24\textwidth}
         \centering
         \includegraphics[width=\textwidth]{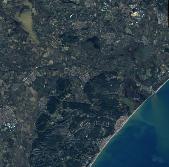}
         \caption{Post-event}
         \label{fig:img2}
     \end{subfigure}
     \hfill
     \begin{subfigure}[b]{0.24\textwidth}
         \centering
         \includegraphics[width=\textwidth]{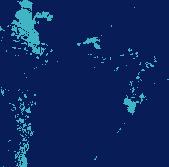}
         \caption{Binary change map}
         \label{fig:img3}
     \end{subfigure}
     \hfill
     \begin{subfigure}[b]{0.24\textwidth}
         \centering
         \includegraphics[width=\textwidth]{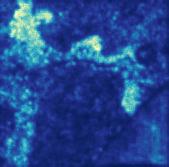}
         \caption{Output map for UDFPN (ResNet-50)}
         \label{fig:img4}
     \end{subfigure}
     
     \caption{Comparison of the change detected by UDFPN vs. the ground truth map for a complex flood event. Inmediate pre-event and post-event images are also depicted for visual analysis.}
     \label{fig:four_images_comparison}
\end{figure}

For the other flood event, a complete visual analysis is depicted at Figure~\ref{fig:four_images_comparison}. In particular, Figure~\ref{fig:img4} shows the ability of our model to generate highly detailed and granular change maps that replicate complex changes in floods. The model is even able to detect false negative pixels in Figure~\ref{fig:img2} around the water terrain (bottom-right corner). 

UDFPN-18 (ResNet-18) offers some improvement in Fires and Floods, but in general, the average performance is worse than other alternatives.

\begin{table}[h]
    \centering
    \begin{tabular}{|l|c|c|c|}
        \hline
        \textbf{Model} & \textbf{GMACs} & \textbf{Parameters (M)} \\ \hline
        UDFPN & 20.63 & 50.41 \\ \hline
        UDFPN-18 & \textbf{11.11} & 25.01 \\ \hline
        Rav\ae{}n & 13.23 (\textbf{0.27}) & \textbf{19.37} \\ \hline
        ResNet-50* & 4.45 & 25 \\ \hline
    \end{tabular}
    \caption{Complexity comparison in terms of MACs (Multiply-Accumulate Operations) and Parameters. Rav\ae{}n multiply its base GMACs (0.27) by total number of patches in 224x224 (49).}
\label{table:efficiency}
\end{table}

Table~\ref{table:efficiency} shows UDFPN-18 is the most efficient alternative even when its detection performance is slightly worse than others. Based on the data provided in the final row of the table, it is evident that the UDFPN-50 encoder is relatively lightweight compared to the entire architecture (UDFPN). This indicates that the computational overhead introduced by the FPN is high.

\section{Discussion}

Following with the experimental analysis, we could assert that our main proposal (UDFPN) offers competitive detection performance on average even when some extreme examples highly downsizing the total AUPRC. The faster alternative (UDFPN-18) should still improve its detection performance. 
In general, experiments are preliminary and more events are demanded for a deeper evaluation. 

From Table~\ref{table:efficiency}, we could assert that UDFPN could perform faster than Rav\ae{}n depending on the version selected. PANN seems to lag behind Rav\ae{}n in time performance as reported in~\cite{smith25}. The model itself is updated during each inference.



To further enhance processing performance, modern space-grade AI accelerators can be utilized as high-efficiency replacements for previous CPU-based implementations (PANN and Rav\ae{}n). Our model is particularly well-suited for these hardware architectures, as its design prioritizes intensive matrix computations and optimized single-forward passes. 

According to~\cite{sang26}, our model’s requirements align closely with the capabilities of leading AI platforms, such as the NVIDIA Jetson family and Xilinx FPGA devices. These mid-power solutions offer robust resource budgets, typically providing computing throughput exceeding 10 TOPS (post-quantization) and onboard memory capacities surpassing several GBs.

To conclude, the main contributions of this paper are listed as follows:

\begin{itemize}
    \item Model versatility and transferability: beyond providing high-fidelity change maps, the self-supervised RSFM acts as a single unified architectural encoder for multiple orbital applications which significantly optimizes onboard storage and memory management for future space missions. RSFM gracefully facilitate \textbf{global embedding generalization}, leveraging pre-training to ensure competitive performance against ad-hoc models across diverse geographic regions and biomes.
    \item FPN image-level generation: while traditional supervised models require extensive end-to-end training, our results demonstrate that the architectural inductive bias of the FPN is sufficient to re-align deep semantic features. This design choice is not exclusive and could be integrated into other encoders to provide~\textbf{image-level change maps}, thus replacing slow patch-based processing. By skipping training, we significantly streamline the deployment pipeline and risks of final deployment on space hardware (e.g., Vitis AI). 
    \item Efficiency and scalability in extreme environments: 
    The combination of high-resolution and precise change maps (0.25x) and low computational overhead (single pass) makes UDFPN uniquely suited for the current generation of AI-enabled space processors. Future improvements should focus on alleviating computing requirements in FPN.
\end{itemize}

\addcontentsline{toc}{section}{References}
\bibliographystyle{unsrt}
\bibliography{library}

\end{document}